\documentclass[11pt]{article}
\usepackage{amsmath}
\usepackage{amssymb}
\usepackage{amsthm}
\usepackage{graphicx}
\usepackage{bm}
\usepackage{breqn}
\usepackage{booktabs}
\usepackage[margin=1in,headheight=25pt]{geometry}
\newtheorem{theorem}{Theorem}

\theoremstyle{definition}

\usepackage{mathtools}
\usepackage{url}
\usepackage{algorithmicx}
\usepackage{algpseudocode}
\usepackage{algorithm}
\DeclareMathOperator*{\E}{\mathbb{E}}

\newcommand{\cvar}{\text{CVaR}_\alpha}
\newcommand{\var}{\text{VaR}_\alpha}

\title{Thesis Proposal: Algorithmic Social Intervention}
\author{Bryan Wilder\\Department of Computer Science and Center for Artificial Intelligence in Society\\University of Southern California\\bwilder@usc.edu}
\date{}

\begin{document}
	\maketitle
	
	\begin{abstract}
		Social and behavioral interventions are a critical tool for governments and communities to tackle deep-rooted societal challenges such as homelessness, disease, and poverty. However, real-world interventions are almost always plagued by limited resources and limited data, which creates a computational challenge: how can we use algorithmic techniques to enhance the targeting and delivery of social and behavioral interventions? The goal of my thesis is to provide a unified study of such questions, collectively considered under the name ``algorithmic social intervention". This proposal introduces algorithmic social intervention as a distinct area with characteristic technical challenges, presents my published research in the context of these challenges, and outlines open problems for future work. A common technical theme is decision making under uncertainty: how can we find actions which will impact a social system in desirable ways under limitations of knowledge and resources? The primary application area for my work thus far is public health, e.g.\ HIV or tuberculosis prevention. For instance, I have developed a series of algorithms which optimize social network interventions for HIV prevention. Two of these algorithms have been pilot-tested in collaboration with LA-area service providers for homeless youth, with preliminary results showing substantial improvement over status-quo approaches. My work also spans other topics in infectious disease prevention and underlying algorithmic questions in robust and risk-aware submodular optimization.
	\end{abstract}
	
	\section{Introduction} 
	
	My research examines how techniques in artificial intelligence (including optimization, machine learning, game theory, and social network analysis) can be used to intervene in social and behavioral systems. Societies around the world face challenges of enormous scale: preventing and treating disease, confronting poverty, and a range of other issues impacting billions of people. In response, governments and communities deploy interventions addressing these problems (e.g., outreach campaigns to enroll patients in treatment or educational programs to raise awareness about preventative strategies). However, these interventions are always subject to limited resources and are deployed under considerable uncertainty about properties of the system; deciding manually on the best way to deploy an intervention is extremely difficult. 
	
\begin{figure}
		\centering
		\includegraphics[width = 6in]{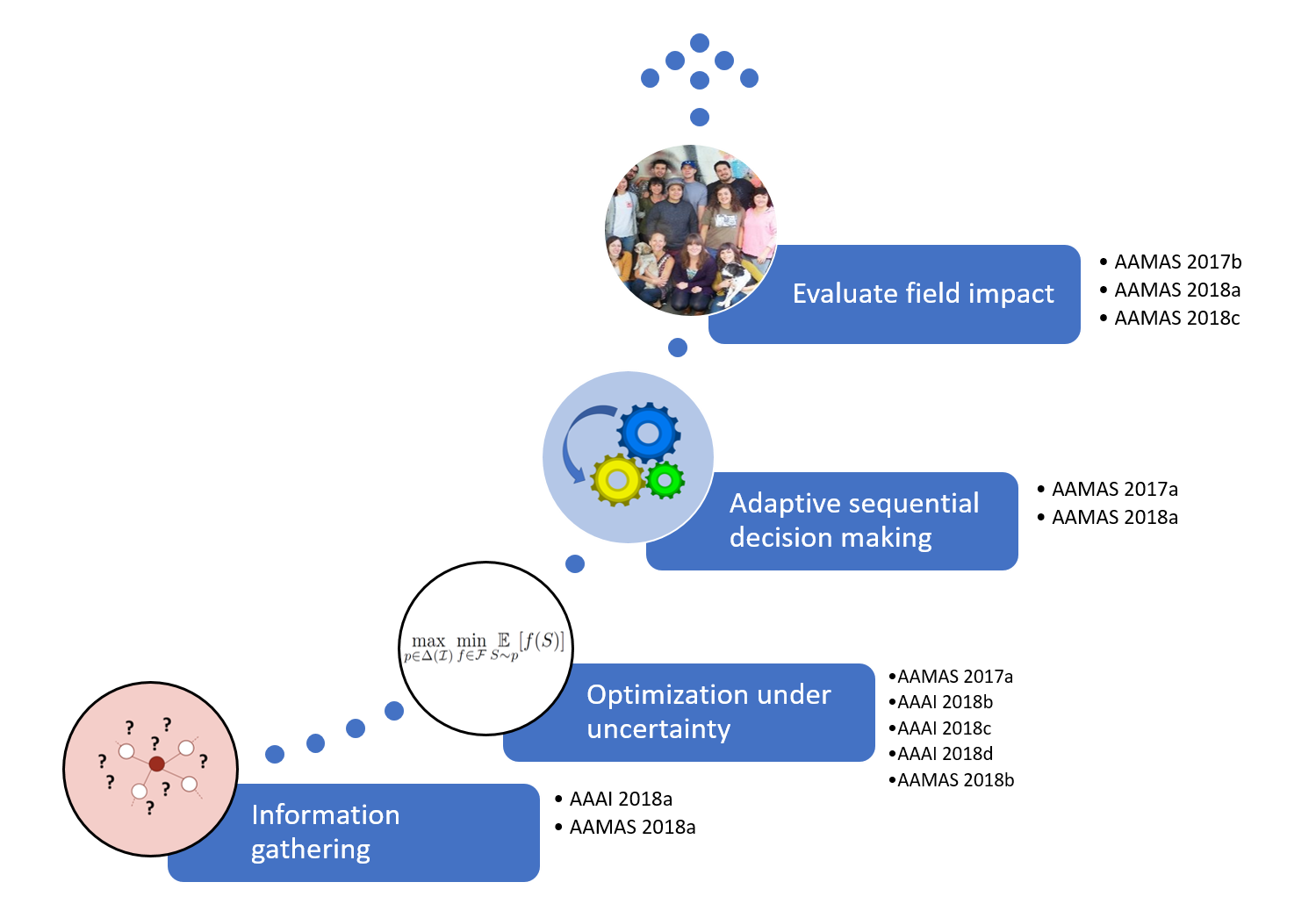}
		\caption{Technical components of algorithmic social intervention and related publications.} \label{fig:research_themes}
	\end{figure}

	Motivated by such challenges, the goal of this thesis is to establish a set of algorithmic techniques which confront underlying challenges in the delivery of social and behavioral interventions (across both public health and other areas) and to field-test these techniques in socially impactful problem settings. We refer to this domain as \emph{algorithmic social intervention}. Social intervention domains motivate a range of common technical challenges (see Figure \ref{fig:research_themes} and Section \ref{section:asi} for more details). My published work spans all of these areas, though many interesting open problems remain. Specifically, I have studied information gathering \cite{wilder2018maximizing,wilder2018end}, optimization under uncertainty \cite{wilder2017uncharted,wilder2018equilibrium,wilder2018risk,wilder2018preventing,wilder2018optimizing} and adaptive sequential decision making \cite{wilder2017uncharted,wilder2018end}. Together with partners in social work and nonprofit agencies, I have empirically evaluated two of the resulting algorithms, with pilot tests showing substantial improvements over status-quo techniques \cite{yadav2017influence,wilder2018end}. 
	
	Specifically, my research thus far has focused on algorithmic approaches to target and enhance interventions in public health settings. One line of work focuses on HIV prevention among homeless youth, where information about HIV is spread through the youths' social network. The challenge is selecting influential peer leaders who will be able to maximize the spread of the resulting diffusion. I have developed a set of algorithms for selecting peer leaders under uncertainty about the structure of the social network and how information propagates \cite{wilder2017unknown,wilder2018maximizing,wilder2018end}. Two of these algorithms have been pilot-tested with LA-area drop in centers serving homeless youth. These studies show substantial improvement over the status-quo heuristic used to select peer leaders \cite{wilder2018end,yadav2017influence}.  Another area is infectious disease prevention, where the challenge is to target limited intervention resources (e.g., outreach campaigns to improve treatment uptake) to the population groups which will have the largest impact on overall disease rates. I developed an algorithm to near-optimally target such interventions, with a particular focus on the problem of reducing tuberculosis spread in India \cite{wilder2018preventing}. In simulation, this algorithm averts over 8,000 cases of tuberculosis per year compared to the status-quo policy.
	Underlying these applications are a number of fundamental technical challenges, related to decision making under uncertainty. Endemic to public health interventions is a lack of information about the system: where the problems lie, how agents interact, and ultimately what outcome an intervention will have. Similar challenges arise in social and behavioral interventions across numerous contexts. Much of my work formalizes underlying challenges in decision making under uncertainty which are motivated by such applications, develops algorithmic solutions, and proves theoretical guarantees on their performance. The ultimate objective is algorithms that come with both rigorous theoretical analysis and field-tested practical performance. Towards this end, I have developed algorithms for \emph{robust} \cite{wilder2018equilibrium} and \emph{risk-averse} \cite{wilder2018risk} submodular optimization. Submodularity formalizes a natural diminishing returns property which occurs across many settings (including the HIV and tuberculosis prevention settings discussed above), making submodular optimization under uncertainty an important and natural algorithmic challenge.
		
	The remainder of this proposal is organized as follows. Section \ref{section:asi} defines the area of algorithmic social intervention in greater detail and expands on the technical challenges common in such domains. Sections \ref{section:im}, \ref{section:submod}, and \ref{section:disease} survey completed research, divided by focus area: Section \ref{section:im} covers influence maximization in the field, Section \ref{section:submod} covers submodular optimization under uncertainty, and Section \ref{section:disease} covers infectious disease prevention. Lastly, Section \ref{section:future} discusses proposed future work. 
	
	\section{Algorithmic Social Intervention}\label{section:asi}

	The goal for this thesis is to establish a unified study of algorithmic social intervention: computational approaches for optimally targeting and enhancing social and behavioral interventions to achieve policy or community-level goals. The aim is to bridge algorithm design, optimization, and machine learning with practice, field deployments, and social impact. Relevant domains are often characterized by the following goals and challenges (though not all may be present in a single domain):
	
	\begin{itemize}
		\item Interventions are delivered in a preexisting social context composed of many agents with their own goals and behaviors. The interactions of these agents collectively produce the system's behavior. 
		\item Agents' behaviors are not totally determined by the intervention: particular incentives, services, or rules may be introduced, but then agents make their own decisions in response.
		\item Agents are not perfectly rational, requiring the use of models and techniques from the social and behavioral sciences to describe behavior.
		\item There are many unknowns: the dynamics and interactions between agents are complex and are not fully specified by the available data.
		\item Applications often focus on vulnerable or underserved populations. 
		
	\end{itemize}

	Figure \ref{fig:research_themes} divides the underlying technical challenges of such domains into several stages. Each stage also lists associated publications. The first stage is \emph{information gathering}. Here, the challenge is to acquire the data needed to optimize the intervention in an efficient manner. For instance, in a social network intervention, it may be necessary to minimize the number of nodes who are surveyed to obtain edges. The second stage is \emph{optimization under uncertainty}. Since the available data is rarely enough to fully specify the objective function, methods such as robust, stochastic, or risk-aware optimization are necessary. The third stage is \emph{adaptive sequential decision making}. Once an intervention is in progress, the algorithmic system has the opportunity to interact with the world, observe the consequences of its decisions, and adjust accordingly. Lastly, a critical part of algorithmic social intervention is to \emph{evaluate field impact}. While typical means of assessment (theoretical analysis, simulation experiments) are important tools, it is critical to validate the algorithm in a field experiment, ideally in comparison to alternate heuristics and algorithms. 

	\section{Influence maximization in the field} \label{section:im}
	
	Influence maximization is a crucial technique used in preventative health interventions, such as HIV prevention amongst homeless youth. Drop-in centers for homeless youth train a subset of youth as peer leaders who will disseminate information about HIV through their social networks. The challenge is to find a small set of peer leaders who will have the greatest possible influence. While many previous algorithms have been proposed for influence maximization \cite{kempe2003maximizing,chen2010scalable,jung2012irie,tang2014influence}, none fully address the challenges of influence maximization in a field setting. Across public health (and other) settings, agencies will be uncertain about the structure of the social network and how information propagates. Accordingly, it is necessary to develop algorithms which gather only the most parsimonious amount of information required to locate influential seeds and incorporate remaining unknowns into the optimization process. Moreoever, practical algorithms must also handle the contingencies of real-world deployments; for instance, youth invited to attend an intervention may simply fail to show up. This line of work develops a series of influence maximization algorithms which address such challenges.
	
	\subsection{Algorithms}
	
	\subsubsection{DOSIM: robust optimization under parameter uncertainty}
	The DOSIM algorithm, developed in \cite{wilder2017unknown} performs robust optimization under uncertainty about how influence spreads through the social network. We first formalize the influence maximization problem as follows. The youth have a social network represented as a graph $G = (V, E)$. Each youth is initially inactive, meaning that they have not received information about HIV prevention. Once nodes are activated by the intervention, they have a chance to influence their peers. We model this process through a variant on the classical independent cascade model (ICM) which has been used by previous work on HIV prevention  and better reflects realistic time dynamics \cite{yadav2016using,wilder2017uncharted,yadav2017influence}. The process unfolds over discrete time steps $t = 1...T$, where $T$ is a time horizon. There is a propagation probability $p_e$ for each edge $e$. When a node becomes active, it attempts to activate each of its neighbors. Each attempt succeeds independently with probability $p_e$. Activation attempts are made at each time step until either the neighbor is influenced or the time horizon is reached. The objective is to select a set of $K$ seed nodes at each time step $t$ so that the expected total influence spread is maximized. 
	
	The key challenge is that the propagation probabilities $p_e$ are not known. We model this as a zero-sum game between the influencer, who selects the seed nodes, and an adversary (nature) who selects the true $p_e$. The goal of the influencer is to find a strategy which performs near-optimally regardless of the unknown parameters, that is, their payoff in the game is the ratio of the expected influence spread resulting from their chosen seed set to the optimal influence spread achievable if the true parameters were known in advance. 
	
	The algorithmic challenge is to compute equilibria in this game. However, it is not apparent how to do so, since both players have extremely large strategy spaces. In particular, nature has an \emph{infinite} strategy space consisting of all (continuous) choices for the unknown parameters subject to interval uncertainty, while the influencer can choose from all possible subsets of seed nodes. In order to resolve this dilemma, two key technical approaches are used in \cite{wilder2017uncharted}. First, to handle nature's infinite strategy space, it is proved that the strategy space may be discretized to a polynomial number of points with only arbitrarily small loss. Second, to handle the influencer's exponentially large number of actions, a \emph{double oracle} approach is employed. Double oracle is an approach for solving large zero-sum games which incrementally builds an equilibrium starting from a small number of strategies. The algorithm proceeds over a series of iterations. At the first iteration, each player is restricted to a small number of pure strategies. We compute a minimax equilibrium in this restricted game (e.g., via linear programming) and then find each player's (approximate) best response to current mixed strategy of their opponent. This best response is added to the player's current strategy set, and the algorithm continues to iterate. Convergence to an equilibrium is guaranteed when the best response of each player is already contained in their current strategy set. In \cite{wilder2017uncharted}, we show in simulation that DOSIM results in substantially more robust solutions that simply planning based on a set of nominal parameters. In Section \ref{section:field}, we also give field results from \cite{yadav2017influence} showing that DOSIM is empirically successful at finding high-quality seed sets in a real world pilot study. 
	
	\subsubsection{ARISEN: influence maximization with an unknown network}
	
	Previous algorithms for influence maximization assume that the social network is given explicitly as input. However, in many real-world domains, the network is not initially known and must be gathered via laborious field observations. For example, collecting network data from vulnerable populations such as homeless youth, while  crucial for health interventions, requires significant time spent gathering field observations \cite{rice2012mobilizing}. Social media data is often unavailable when access to technology is limited, for instance in developing countries or with vulnerable populations. Even when such data is available, it often includes many weak links which are not effective at spreading influence \cite{bond201261}. For instance, a person may have hundreds of Facebook friends whom they barely know. In principle, the entire network could be reconstructed via surveys, and then existing influence maximization algorithms applied.  However, exhaustive surveys are very labor-intensive and often considered impractical \cite{valente2007identifying}. For influence maximization to be relevant to many real-world problems, it must contend with limited \emph{information} about the network, not just limited \emph{computation}.
	
	The major informational restriction is the number of nodes which may be surveyed to explore the network. Thus, a key question is: \emph{how can we find influential nodes with a small number of queries?} We formalize this problem as \emph{exploratory influence maximization} and seek a principled algorithmic solution, i.e., an algorithm which makes a small number of queries and returns a set of seed nodes which are approximately as influential as the globally optimal seed set. Each query targets a given node in the graph and reveals all of that node's edges. At each step, the algorithm may either query the neighbor of a previously queried node, or query a uniformly random node in the graph. Existing field work uses heuristics, such as sampling some percentage of the nodes and asking them to nominate influencers \cite{valente2007identifying}. To our knowledge, no previous work directly addresses this question from an algorithmic perspective.
	
	We show that for general graphs, any algorithm for exploratory influence maximization may perform arbitrarily badly unless it examines almost the entire network. However, real world networks often have strong \emph{community} structure, where nodes form tightly connected subgroups which are only weakly connected to the rest of the network \cite{leskovec2009community}. Consequently, influence mostly propagates locally. Community structure has been used to develop computationally efficient influence maximization algorithms \cite{wang2010community,chen2014cim}. Here, we use it to design a highly information-efficient algorithm. We make four contributions. \emph{First}, we introduce exploratory influence maximization and show that it is intractable for general graphs. \emph{Second}, we present the ARISEN algorithm, which exploits community structure to find influential nodes. \emph{Third}, we show that ARISEN has strong empirical performance on an array of real world social networks. \emph{Fourth}, we formally analyze ARISEN on graphs drawn from the Stochastic Block Model (SBM) \cite{fienberg1981categorical}, a widely studied model of community structure. We prove that it approximates the optimal influence if the entire network were known by querying only a polylogarithmic number of nodes in the network size.
	
	We give the main idea behind the algorithm here and defer a formal description to the main paper \cite{wilder2018maximizing}. In a graph with community structure, a reasonable strategy is to try and select one seed node from each community (or the $K$ largest communities if we have a budget of $K$ seed nodes). The rationale is that we expect influence to propagate widely within a given community but only to a limited extent between communities. So, multiple seed nodes within a given community would be redundant compared to seeding another community entirely. The goal of the ARISEN algorithm is to use a small number of queries to choose seed nodes which are likely to lie in the $K$ largest communities. The underlying approach is as follows. First, we sample a set of prospective seed nodes uniformly at random. Then, we use queries to simulate a random walk around each prospective seed; it can be shown that this random walk will stay (with high probability) within the starting community. The nodes encountered on this random walk are used to estimate the average degree of the community, which is in turn used to estimate the community's size. Using these estimates, ARISEN constructs a probability distribution over the prospective seeds and samples each actual seed independently at random from this distribution. The main challenge is that we cannot in general tell whether two prospective seed nodes lie in the same community, and so we must construct a distribution which \emph{implicitly} leads to seed nodes in different communities being selected. The number of times a given community is sampled as a prospective seed is proportional to that community's size. Hence, ARISEN's probability distribution assigns each prospective seed node weight \emph{inversely} proportional to its community's estimated size. This evens out the sampling bias towards large communities and ensures that (in expectation) each of the largest $K$ communities is seeded exactly once. 
	
	We analyze ARISEN theoretically on graphs which are drawn from the Stochastic Block Model (SBM), a common model of community structure. The SBM originated in sociology \cite{fienberg1981categorical} and lately has been intensively studied in computer science and statistics (see e.g.\ \cite{abbe2015community,krzakala2013spectral,mossel2015reconstruction}). In the SBM, the network is partitioned into disjoint communities $C_1....C_L$. Each within-community edge is present independently with probability $p_w$ and each between-community edge is present independently with probability $p_b$. Recall that the Erd\H{o}s-R\'enyi random graph $\mathcal{G}(n, p)$ is the graph on $n$ nodes where every edge is independently present with probability $p$. In the SBM, community $C_i$ is internally drawn as $\mathcal{G}(|C_i|, p_w)$ with additional random edges to other communities. While the SBM is a simplified model, our experimental results show that ARISEN also performs well on real-world graphs. ARISEN takes as input the parameters $n, p_w$, and $p_b$, but is not given any prior information about the realized draw of the network. It is reasonable to assume that the model parameters are known since they can be estimated using existing network data from a similar population (in our experiments, we show that this approach works well). For instance in HIV prevention, homeless youth social networks have been shown to exhibit community structure and several studies have gathered networks from which to infer $p_w$ and $p_b$ \cite{yadav2016using,rice2012mobilizing}. 
	
	We state here a simplified version of our main theoretical result which captures the intuition. Suppose that the top $K$ communities each have equal size $\mu$, and occupy a linear portion of the network -- for concreteness, $\mu K \geq 0.01 n$. We have
	
	\begin{theorem}[Simplified case]
		Under the above conditions, ARISEN can be implemented with approximation ratio $(1 - \frac{1}{e^{0.99}} - 0.01 - \frac{1}{K} - o(1)) \beta(\mu)$ using $O(\log^6 n)$ queries. \label{theorem:simple}
	\end{theorem}
	
	Here, $\beta(\mu)$ is a constant which depends on $p_w$ and $q$. We a defer a detailed explanation to the paper \cite{wilder2018maximizing} and just note here that $\beta(\mu)$ is the fraction of nodes contained in the giant connected component of an Erd\H{o}s-R\'enyi random graph $\mathcal{G}(\mu K, p_w\cdot q)$. The query cost is chosen so that the random walk based estimates of each community's size are accurate with high probability. We emphasize that only a polylogarithmic number of nodes need be queried, an exponential improvement over exhaustive surveys. The first term in the approximation ratio is nearly $1 - 1/e$, up to error terms which decrease as $n$ and $K$ become large. We show that each of the top $K$ communities is seeded with probability close to $1 - 1/e$. The second term, $\beta(\mu)$, is the fraction of each of the top $K$ communities which can be influenced by a seed node. 
	
	In the full paper, simulation results bear out the main conclusion, that ARISEN is able to find influential seed nodes with a small number of queries. Experimentally, ARISEN outperforms a range of heuristics and is often able to closely approximate the optimal influence spread while querying 15-20\% of the network.

	\subsubsection{CHANGE: end-to-end influence maximization for the field}
	
	Thus far, algorithms for influence maximization in the field have a high barrier to entry: they require a great deal of time to gather the complete social network of the youth, expertise to select appropriate parameters, and computational power to run the algorithms. None of these are likely available to resource-strained service providers who will ultimately be the ones to deploy influence maximization. This paper \cite{wilder2018end} presents CHANGE, a novel system for influence maximization to ameliorate these difficulties. CHANGE draws on the insights used to develop DOSIM and ARISEN, but is tailored to the constraints of a field deployment in public health settings. Specifically, CHANGE is designed to avoid DOSIM's high computational cost and circumvent some practical difficulties in deploying ARISEN. Specifically, it is difficult to use ARISEN's random walk based procedure with homeless youth, because it is often infeasible to locate a sequence of youth at the agency (youth may not be at the agency that day or be otherwise unreachable). Hence, CHANGE should be thought of as a streamlined, field-ready system which draws on a series of insights into influence maximization among homeless youth but which lacks theoretical guarantees for some components of the system. 
		
	CHANGE is easy to deploy, but this simplicity is crucially enabled by a series of insights into the social structure of homeless youth (which may be useful for other vulnerable populations). We conducted a pilot test of CHANGE's performance in a real deployment by a drop-in center serving homeless youth in a major U.S.\ city. CHANGE was used to plan a series of interventions designed to spread HIV awareness among the youth. \emph{CHANGE obtained comparable influence spread to state of the art algorithms while surveying only 18\% of nodes for network data}, a finding which is backed by additional simulation results.
	
	Overall, CHANGE offers a practical, field-tested vehicle for deployed influence maximization which drastically lowers the barrier to entry. \emph{To our knowledge, this is the first real-world pilot study of a network sampling algorithm for influence maximization and only the second ever field test of \emph{any} influence maximization algorithm.}
	
\begin{figure}
		\centering
		\includegraphics[width=4in]{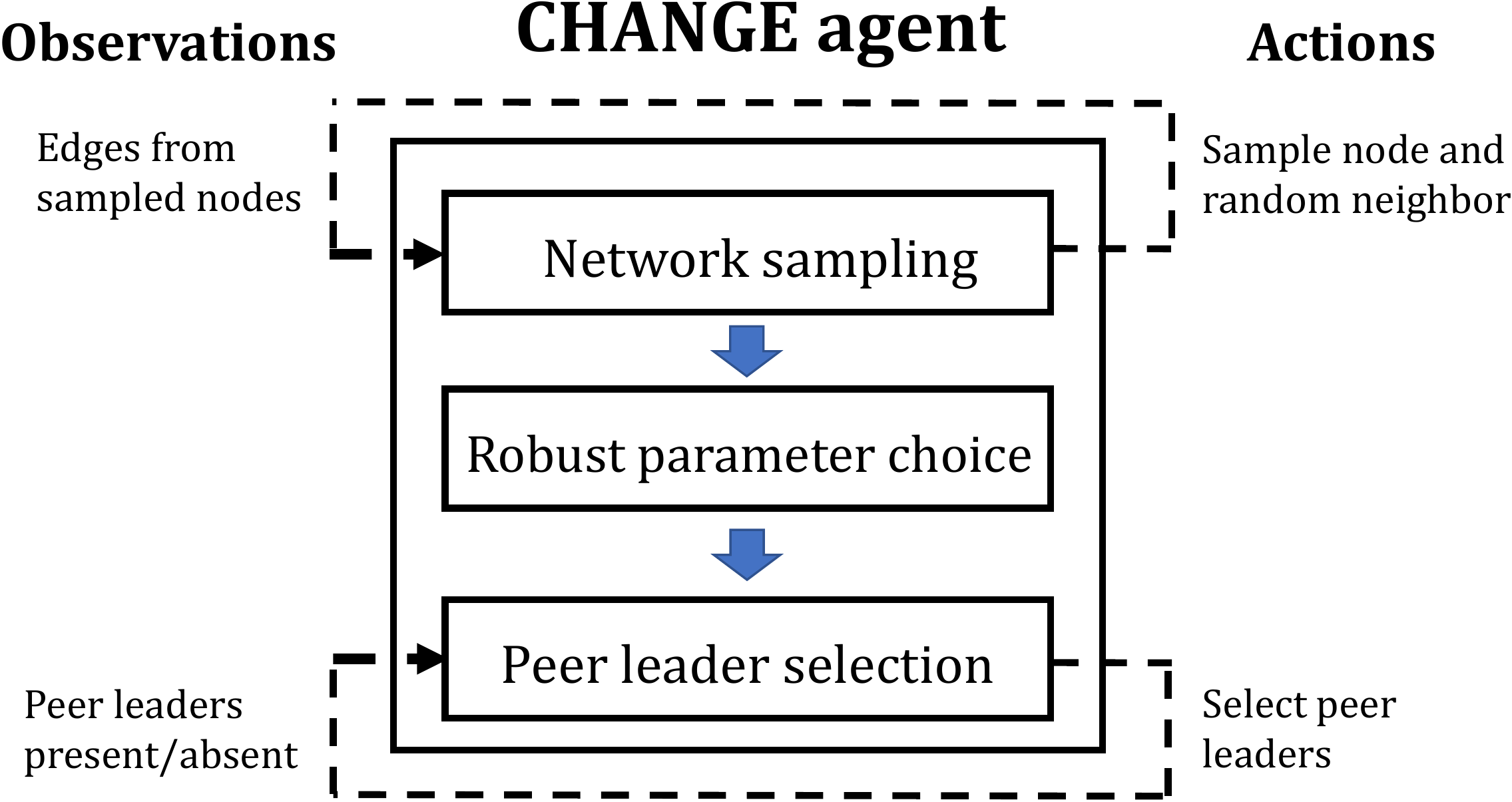}
		\caption{The CHANGE agent.}\label{fig:change}
	\end{figure}

	\textbf{Overview of algorithmic contributions: } We now summarize how CHANGE handles the challenges above. A diagram of the agent can be found in Figure \ref{fig:change}. 
	
	First, to address the \emph{data gathering} challenge, we present an easily deployable sampling protocol which randomly selects a small set of youth to interview. For each of these youth, a randomly chosen neighbor is also interviewed. We show that this procedure gathers enough of the network to enable high-quality influence maximization even though it surveys only a small number of nodes directly.
	
	Second, to address \emph{computational power} challenge (which in turn stems from unknown parameters), we present a heuristic for selecting influence maximization solutions which are robust to uncertainty in the probability $p$ that influence will spread. We show that this heuristic finds solutions which obtain approximately 90\% of the maximum possible influence spread under \emph{any} value for $p$. Importantly, this heuristic runs in minutes on a laptop, while DOSIM (the previously proposed algorithm for this problem) requires hours or even days of time on a high performance computing cluster. 
	
	Third, we integrate these components with an \emph{adaptive greedy} algorithm for planning interventions and prove the first theoretical guarantee for influence maximization under execution errors. The challenge is that some youth selected as peer leaders may not attend the intervention \cite{wilder2017uncharted,yadav2017influence}. Our algorithm selects its action with such uncertainties in mind, observes which youth do attend, and then plans the next round using this observation. We prove that it obtains a constant-factor approximation to the \emph{optimal} adaptive policy.
	
	A detailed presentation of the CHANGE agent can be found in the full paper \cite{wilder2018end}. However, the next section presents field results for both DOSIM and CHANGE from pilot studies carried out in collaboration with LA-area drop in centers serving homeless youth.  
	
	\subsection{Pilot tests and field results} \label{section:field}
	
	In the pilot tests, trained social workers delivered the \emph{Have You Heard} intervention, previously published in the public health literature \cite{rice2012mobilizing}. The social workers conducted a day-long class with the selected youth, covering HIV awareness and prevention, and training the youth as peer leaders to communicate with other youth at the agency. Four pilot tests have been conducted so far, each using a different algorithm to select the peer leaders. Each pilot test used a distinct population with its own social network. The four algorithms were DOSIM and CHANGE, introduced above, along with HEALER \cite{yadav2016using} (a previously developed algorithm for the probem) and degree centrality (DC). DC is the status-quo heuristic used by agencies, and simply picks the highest degree nodes to seed. 
	
	CHANGE first queried a subset of 18\% of nodes for network data, while DOSIM, HEALER, and DC received the full network in advance. Three sets of peer leaders were selected by each algorithm, with approximately 4 peer leaders in each set. Peer leaders were paid \$60.  One month after the start of the study, we conducted a follow up survey with all of the youth who initially enrolled. We asked the youth whether they had received information about HIV prevention from a peer who was part of the study. Youth were paid \$20 to respond to the follow up survey. The field results are reported across two separate papers (\cite{yadav2017influence} for HEALER, DOSIM, and DC, and \cite{wilder2018end} for CHANGE), but both studies used identical protocols. 60-70 participants were recruited for each study.
	
\begin{figure}
		\centering
		\includegraphics[width=3in]{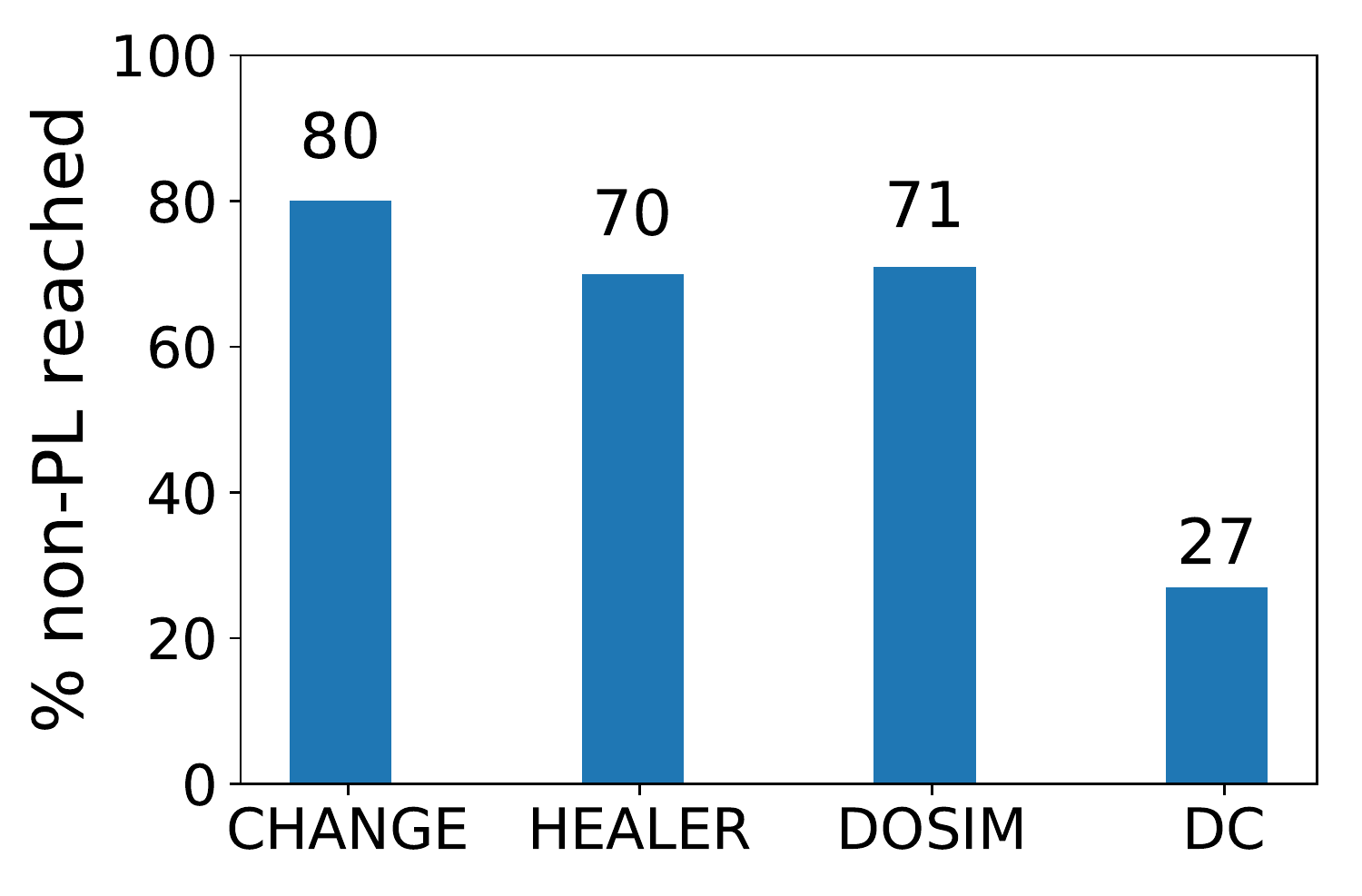}
		\caption{Percentage of non-peer leaders who reported receiving information about HIV in the pilot study corresponding to each algorithm.} \label{fig:pilot}
	\end{figure}

	Figure \ref{fig:pilot} presents the main result: the amount of influence spread generated by each algorithm. Specifically, we used the follow up survey to examine the percentage of youth who were not peer leaders who reported that they received information about HIV prevention. We see that the AI-based algorithms (CHANGE, HEALER, DOSIM) do fairly well, reaching 70-80\% of non-peer leaders. However, DC performs poorly, reaching only 27\% of non-peer leaders. While these results are preliminary, they show that there is promise in using algorithmic techniques to enhance influence maximization interventions. We also note that CHANGE performs just as well as HEALER and DOSIM despite querying only 18\% of the network for links. We do not claim that CHANGE actually outperforms the other two algorithms (the difference could be caused by small sample sizes or other external factors); however, the close results indicate that it may be possible to find influential nodes using only a limited amount of network data. More detailed analysis of these results can be found in \cite{yadav2017influence,wilder2018end}, where we examine the robustness of these results through simulation, give more detailed explanations for the differences between algorithms, and formulate general insights and lessons learned about influence maximization in a field setting.

	\section{Submodular optimization under uncertainty} \label{section:submod}
	
	Inspired by the challenges of influence maximization with limited data, this next section considers the general problem of optimizing a monotone submodular function under uncertainty about the true objective. We study the problem from two perspectives (presented in \cite{wilder2018equilibrium} and \cite{wilder2018risk} respectively). First, \emph{robust optimization}, where the goal is to maximize the worst case from a set of possible objectives. Second, \emph{risk-averse optimization}, where we aim to avoid disastrous outcomes instead of simply maximizing expected utility. In both cases, we substantially improve the existing state of the art, claims that are borne out both by theoretical guarantees and experimental results. 
	
	\subsection{Robust optimization}
	
	Let $X$ be a set of items with $|X| = n$. A function $f: 2^X \to R$ is submodular if for any $A \subseteq B$ and $i \in X\setminus B$, $f(A \cup \{i\}) -f(A) \geq f(B \cup \{i\}) - f(B)$. We restrict our attention to functions that are $\emph{monotone}$, i.e., $f(A \cup \{i\}) - f(A) \geq 0$ for all $i \in X, A \subset X$. Without loss of generality, we assume that $f(\emptyset) = 0$ and hence $f(S) \geq 0 \, \forall S$. Let $\mathcal{I}$ be a collection of subsets of $X$. For instance, we could have $\mathcal{I} = \{S \subseteq X : |S| \leq k\}$. In general, we will allow $\mathcal{I}$ to be any matroid. The objective is to find a utility-maximizing element of $\mathcal{I}$. 
	
	We consider the robust optimization setting where the true objective to be optimized is not known exactly. Instead, it belongs to an \emph{uncertainty set} which gives the set of possibilities consistent with prior knowledge. Let $\mathcal{F} = \{f_1...f_m\}$ be a finite set of submodular functions on the ground set $X$. We are promised that the true objective belongs to $\mathcal{F}$ but do not know which element of $\mathcal{F}$ it is. Accordingly, we aim to maximize the minimum value, $\max_{S \in \mathcal{I}} \min_{f_i \in \mathcal{F}} f_i(S)$. The total number of objective functions $m$ may be very large, potentially exponentially large in the size of the ground set $n$. 
	
	Since the robust submodular optimization problem is in general inapproximable \cite{krause2008robust}, we consider a common relaxation of it to a zero sum game \cite{krause2011randomized,chen2017robust}. We would like to find a minimax equilibrium of the game where the maximizing player's pure strategies are the subsets in $\mathcal{I}$, and the minimizing player's pure strategies are the functions in $\mathcal{F}$. The payoff to the strategies $S \in \mathcal{I}$ and $f_i \in \mathcal{F}$ is $f_i(S)$. We call a game in this form a \emph{submodular best response} (SBR) game. For the maximizing player, computing the minimax equilibrium is equivalent to solving
	\begin{align}\label{problem:randomized}
	\max_{p \in \Delta(\mathcal{I})} \min_{f \in \mathcal{F}} \E_{S \sim p}[f(S)]
	\end{align}
	
	where $\Delta(\mathcal{I})$ is the set of all distributions over the elements of $\mathcal{I}$. Oftentimes, we will work with independent distributions over $X$, which can be fully specified by a vector $\bm{x} \in R^n_+$. $x_i$ gives the marginal probability that item $i$ is chosen. Denote by $p^I_{\bm{x}}$ the independent distribution with marginals $\bm{x}$. 
	
	The equilibrium computation problem has been studied by Krause et al.\ \cite{krause2011randomized} and Chen et al.\ \cite{chen2017robust} using very similar techniques: both iterate dynamics where the adversary plays a no-regret learning algorithm and the decision maker plays a greedy best response. This algorithm maintains a variable for every function in $\mathcal{F}$ and so is only computationally tractable when $\mathcal{F}$ is small. By contrast, we deal with the setting where $\mathcal{F}$ is exponentially large, with the objective function arising from an underlying combinatorial structure. In \cite{wilder2018equilibrium}, we explore two applications falling into this setting: a robust budget allocation problem, and security games played on networks. In both cases, our framework leads to the first sub-exponential time algorithm for the problem. Here, we just state the main algorithmic result. 
	
	We solve Problem \ref{problem:randomized} under the assumption that there is a best response oracle available for the adversary, which computes the minimizing function for a given distribution of the maximizing player. However, we require only a weaker oracle, which we call an \emph{best response to independent distributions oracle} (BRI). A BRI oracle is only required to compute a best response to mixed strategies which are independent distributions, represented as the marginal probability that each item in $X$ appears. Given a vector $\bm{x} \in R_+^n$, where $x_i$ is the probability that element $i \in X$ is chosen, a BRI oracle computes $\arg\min_{f_i \in \mathcal{F}} \E_{S \sim p^I_{\bm{x}}}[f_i(S)]$. We use $S \sim \bm{x}$ to denote that $S$ is drawn from the independent distribution with marginals $\bm{x}$. In some domains (e.g., network security games), a BRI oracle is readily available even when the full best response is NP-hard. 
	
	Our main technical contribution is the EQUATOR algorithm, which computes a $(1 - 1/e)^2$-approximation to Problem \ref{problem:randomized}, modulo an additive loss of $\epsilon$. Crucially, EQUATOR makes only polynomially many calls to the BRI, with no direct runtime dependence on $|\mathcal{F}|$. Specifically, EQUATOR takes time polynomial in $n$, $\frac{1}{\epsilon}$, and $M$, where $M$ is an upper bound on the value of any single item ($M \geq \max_{f_i \in \mathcal{F}, j \in S} f_i(\{j\})$). In general, this results in a pseudopolynomial time algorithm (since there is polynomial dependence on $M$), though $M$ is constant in many cases of interest. 
	
	Since the pure strategy sets can be exponentially large, it is unclear what it even means to compute an equilibrium: representing a mixed strategy may require exponential space. Our solution to this dilemma is to show how to efficiently \emph{sample} pure strategies from an approximate equilibrium mixed strategy. This suffices for the maximizing player to implement their strategy. Alternatively, we can build an approximate mixed strategy with sparse support by drawing a polynomial number of samples and outputing the uniform distribution over the samples. In order to generate these samples, EQUATOR first solves a continuous optimization problem. This continuous relaxation uses the \emph{multilinear relaxations} of the the functions in $\mathcal{F}$ (we refer the reader to \cite{calinescu2011maximizing} for more details on the multilinear relaxation). Essentially, the multilinear extension of a submodular function $f$ defines a continuous function over the hypercube $[0,1]^n$ which agrees with $f$ at the vertices. EQUATOR optimizes the pointwise minimum of the multilinear extensions of the functions in $\mathcal{F}$ and then uses known techniques (see \cite{chekuri2010dependent}) to round the resulting fractional point to a distribution over integral sets. This continuous optimization problem is non-convex and nonsmooth. We design a novel stochastic Frank-Wolfe algorithm which obtains a $(1 - 1/e)$-approximation to the continuous problem. After the rounding step, we have the following guarantee:
	
	\begin{theorem}\label{theorem:main}
		EQUATOR outputs a set $S \in \mathcal{I}$ such that $\min_i \E[f_i(S)] \geq (1 - \frac{1}{e})^2 OPT - \epsilon$ with probability at least $1 - \delta$. Its runtime is  $\mathcal{\tilde{O}}\left(T_1 \frac{M^2 k^2 n}{\epsilon^2} + T_2 \frac{k^4 M^4 n}{\epsilon^4} \log\frac{1}{\delta} \right)$\footnote{The $\mathcal{\tilde{O}}$ notation hides logarithmic terms} where $T_1$ is the time to perform linear optimization over the convex hull of $\mathcal{I}$ and $T_2$ is the time to compute a gradient. 
	\end{theorem}

	We remark that $T_1$ is small ($O(n \log n)$) in cases of interest such as the $k$-cardinality constraint, while $T_2$ is typically dominated by the runtime of the $BRI$. This theoretical result substantially improves over the current state of the art; no-regret learning based algorithms proposed for this problem \cite{krause2011randomized,chen2017robust} work only when $\mathcal{F}$ is small, while the ``double oracle" algorithms often used in practice \cite{jain2013security,jain2011double} may take exponential runtime in the worst case. 
	
\begin{figure*}
		\centering
		\includegraphics[height=1.5in]{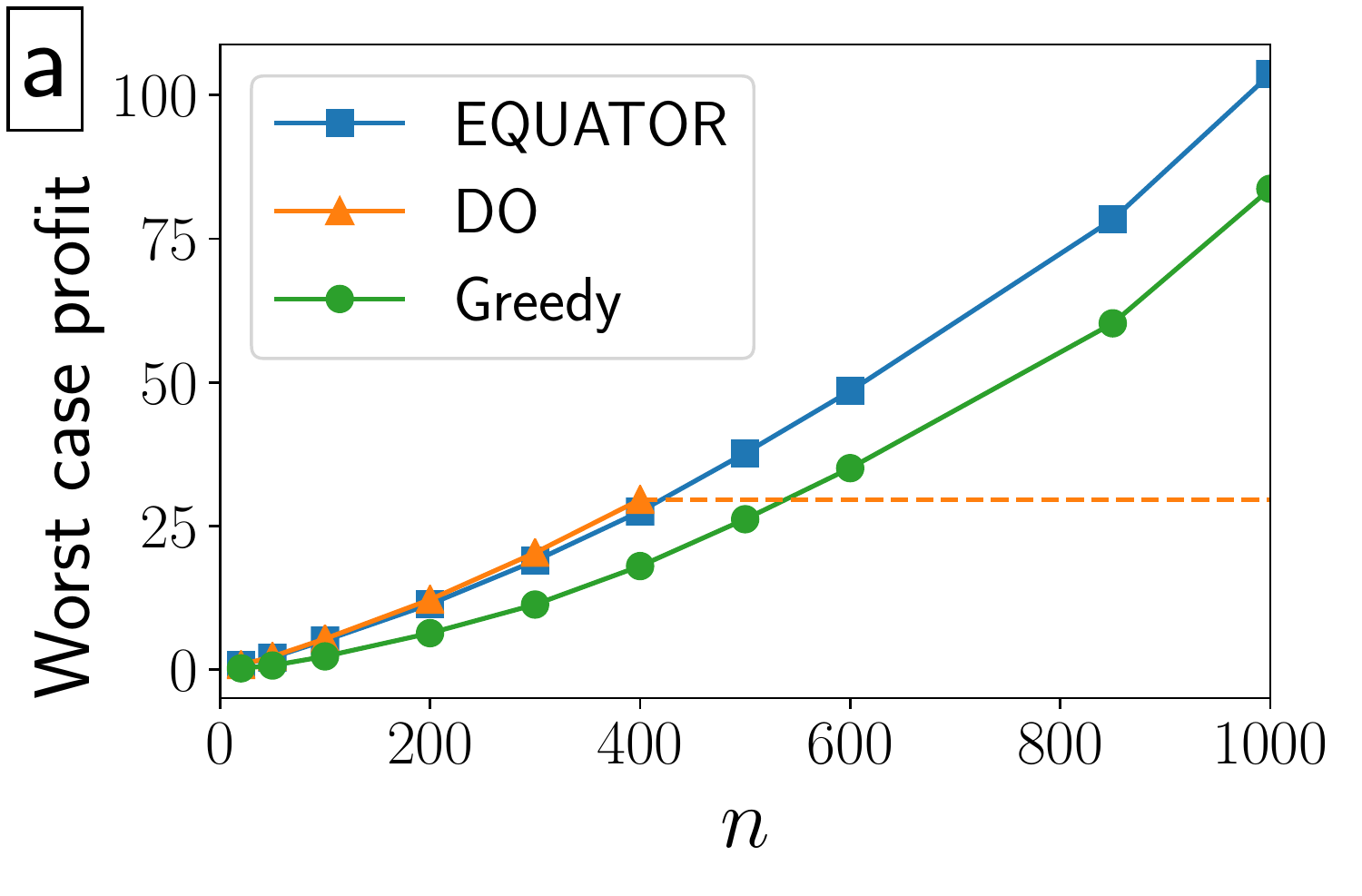}
		\includegraphics[height=1.5in]{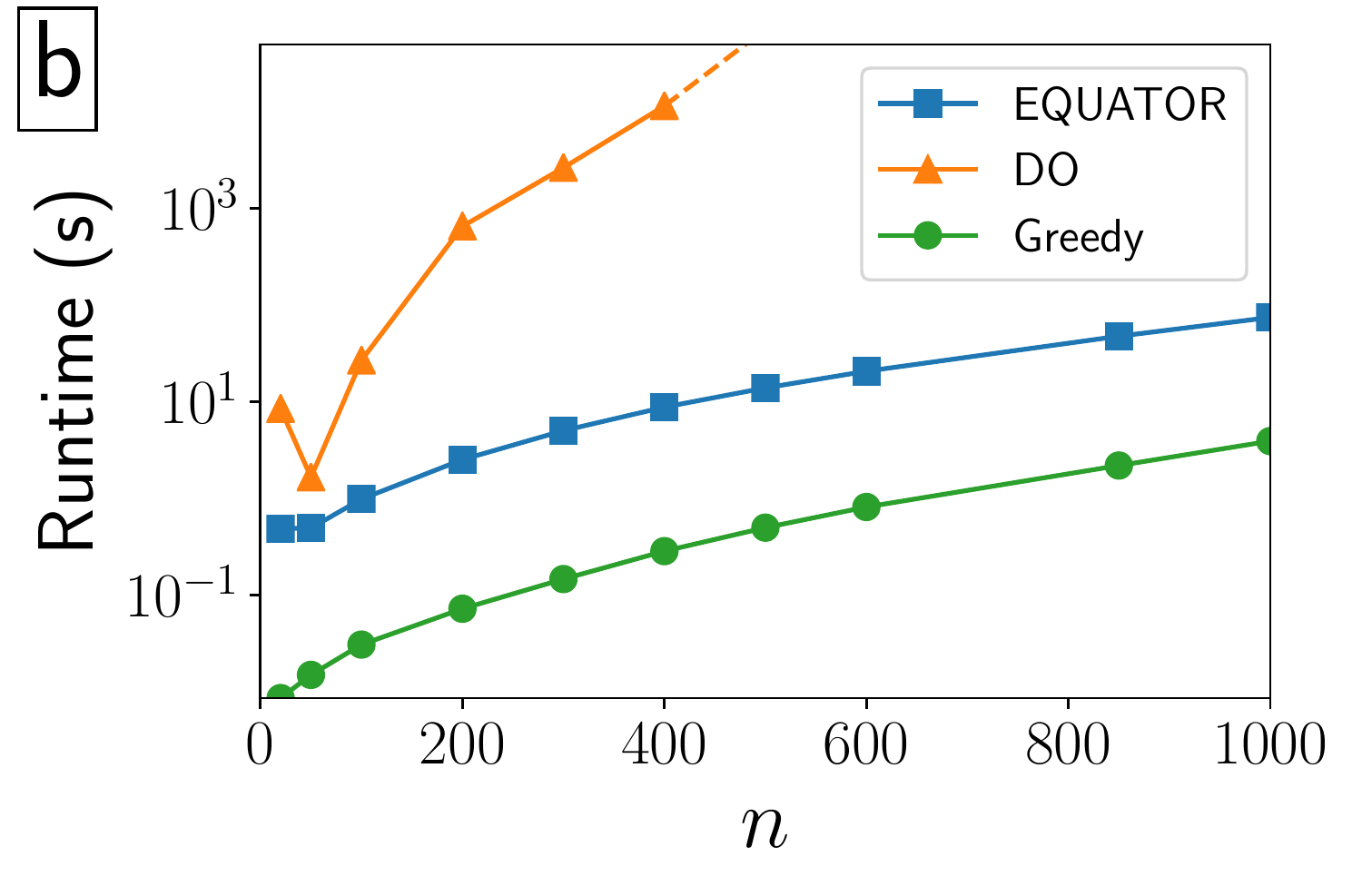}
		\includegraphics[height=1.5in]{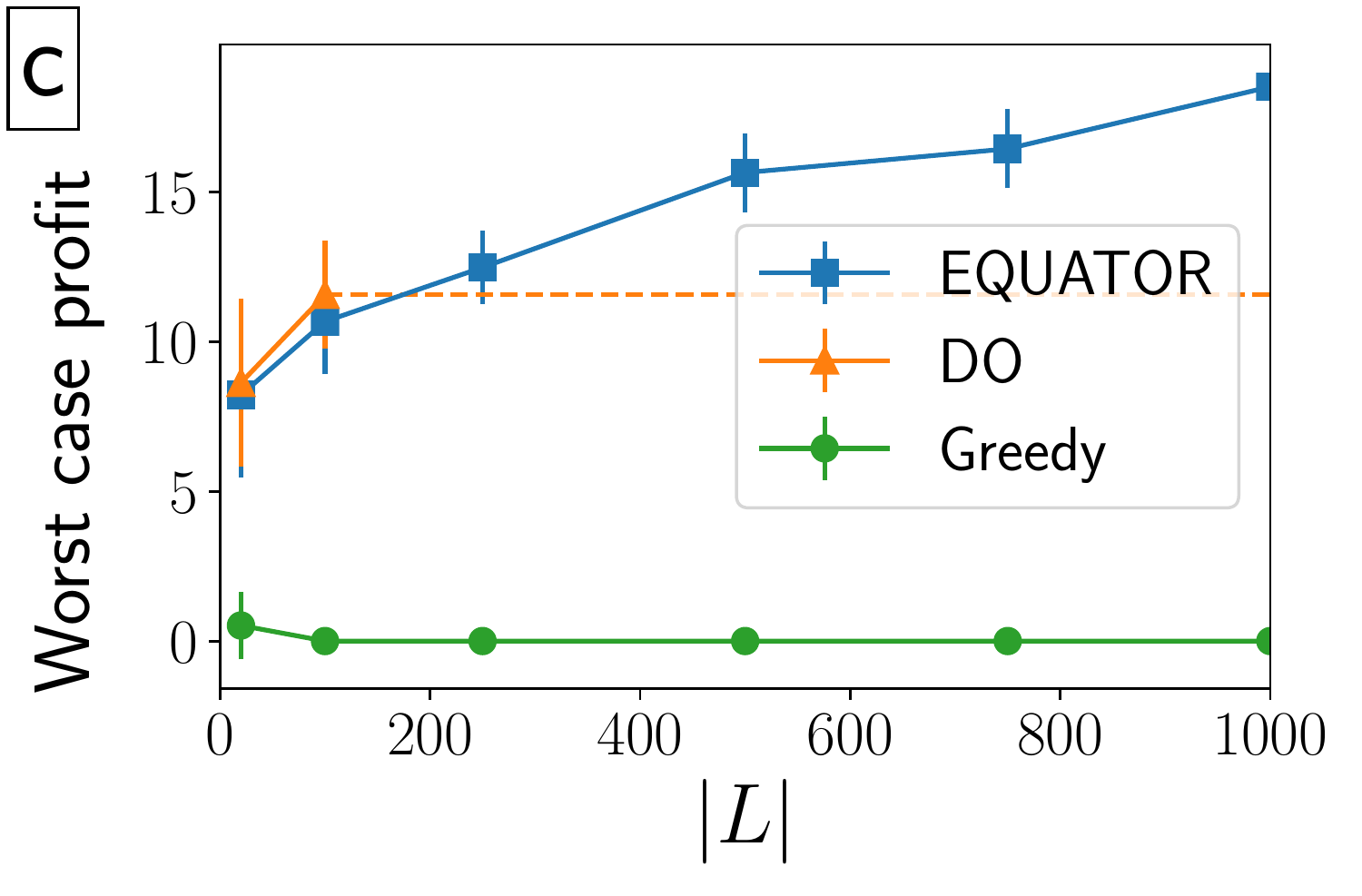}
		\includegraphics[height=1.5in]{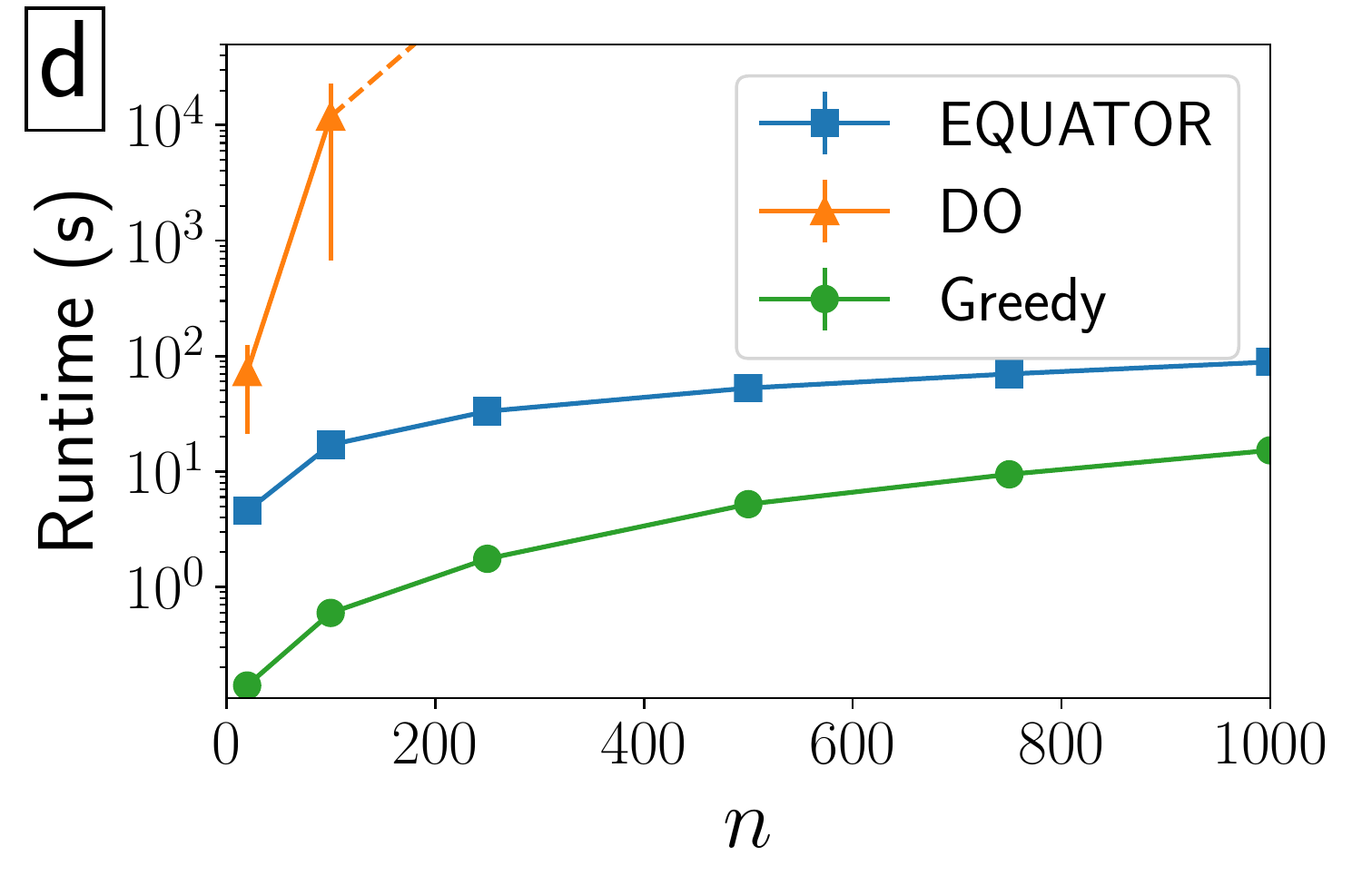}
		\caption{Experimental results for budget allocation.} \label{fig:budget}
	\end{figure*}

	Figure \ref{fig:budget} shows experimental results for a robust budget allocation problem. Budget allocation models an advertiser's choice of how to divide a finite budget $B$ between a set of advertising channels \cite{soma2014optimal,staib2017robust,alon2012optimizing}. Each channel is a vertex on the left hand side $L$ of a bipartite graph. The right hand $R$ consists of customers. Each customer $v \in R$ has a value $w_v$ which is the advertiser's expected profit from reaching $v$. In the robust problem, the profits $\bm{w}$ are not known exactly, instead belonging to an uncertainty set (e.g., based on historical data).

	We compare EQUATOR to the state of the art double oracle algorithm \cite{bosansky2014exact,jain2013security} (DO), which computes a $(1 - 1/e)$-approximate solution but takes exponential time in the worst case. We also compare to a greedy algorithm which optimizes the best-estimate objective function but does not perform any robust optimization. Figure \ref{fig:budget} shows the results; the top row shows randomly generated instances while the bottom row shows results on the Yahoo webscope dataset. Figure \ref{fig:budget}(a) plots the profit obtained by each algorithm when the true $\bm{w}$ is chosen as the worst case in the uncertainty set, with $n$ increasing on the $x$ axis. Figure \ref{fig:budget}(b) plots the average runtime for each $n$. We see that double oracle produces highly robust solutions. However, for even $n = 500$, its execution was halted after 10 hours. Greedy is highly scalable, but produces solutions that are approximately 40\% less robust than double oracle. EQUATOR produces solution quality within 7\% of double oracle and runs in less than 30 seconds with $n = 1000$. In Figure \ref{fig:budget}(c), we see that both double oracle and EQUATOR find highly robust solutions, with EQUATOR's solution value within 8\% of that of double oracle. By contrast, greedy obtains \emph{no} profit in the worst case for $|L| > 20$, validating the importance of robust solutions on real problems. In Figure \ref{fig:budget}(d), we observe that double oracle was terminated after 10 hours for $n = 500$ while EQUATOR scales to $n = 1000$ in under 40 seconds. We conclude that EQUATOR is empirically successful at finding highly robust solutions in an efficient manner, complementing its theoretical guarantees.	
	
	\subsection{Risk-averse optimization}
	
	Decision-making under uncertainty is an ubiquitous problem. Suppose we want to maximize a function $F(\bm{x}, y)$, where $\bm{x}$ is a vector of decision variables and $y$ a random variable drawn from a distribution $D$. A natural approach is to maximize $\E_y\left[F(\bm{x}, y)\right]$, i.e., to maximize the expected value of the chosen decision. However, decision makers are often \emph{risk-averse}: they would rather minimize the chance of having a very low reward than focus purely on the average. This is a rational behavior when failure can have large consequences. For instance, if a corporation suffers a disastrous loss, they may simply go out of business. Or in many cases, low performance entails safety issues. For instance, if a sensor network for water contamination detects problems instantly in 80\% cases, but fails entirely in 20\%, the population will inevitably be exposed to an unacceptable health risk. It is much better to have a sensor network which always detects contaminants, even if it requires somewhat more time on average. 
	
	Hence, it is natural to move beyond average-case analysis and optimize a risk-aware objective function. One widespread choice is the \emph{conditional value at risk} (CVaR). CVaR takes a tunable parameter $\alpha$. Roughly, it measures the performance of a decision in the worst $\alpha$ fraction of scenarios. It is known that when the objective $F$ is a concave function, then CVaR can be optimized via a concave program as well. However, many natural objective functions are \emph{not} concave, and no general algorithms are known for nonconcave functions. We focus on \emph{submodular} functions. Submodularity captures diminishing returns and appears in application domains ranging from viral marketing \cite{kempe2003maximizing}, to machine learning \cite{kulesza2012determinantal}, to auction theory \cite{vondrak2008optimal}. We analyze submodular functions in two settings:
	
	\noindent\textbf{Continuous: } Continuous submodularity, which has lately received increasing attention \cite{bach2015submodular,bian2017guaranteed,staib2017robust} generalizes the notion of a submodular set function to continuous domains. Many well-known discrete problems (e.g., sensor placement, influence maximization, or facility location) admit natural extensions where resources are divided in a continuous manner. Continuous submodular functions have also been extensively studied in economics as a model of diminishing returns or strategic substitutes \cite{kocckesen2000strategic,sampson2016assignment}. Our main result is a $(1 - \frac{1}{e})$-approximation algorithm for maximizing the CVaR of any monotone, continuous submodular function. No algorithm was previously known for this problem.
	
	\noindent\textbf{Portfolio of discrete sets: } Our results for continuous submodular functions also transfer to set functions. We study a setting where the algorithm can select a distribution over feasible sets, which is of interest when the aim is to select a portfolio of sets to hedge against risk \cite{ohsaka2017portfolio}. This is a similar relaxation as in the robust setting studied above. We give a black-box reduction from the discrete portfolio problem to CVaR optimization of continuous submodular functions, allowing us to apply our algorithm for the continuous problem. The state of the art for the discrete portfolio setting is an algorithm by Ohsaka and Yoshida \cite{ohsaka2017portfolio} for CVaR influence maximization. Our results are stronger in two ways: (i) they apply to \emph{any} submodular function and (ii) give stronger approximation guarantee. Allowing the algorithm to select a convex combination of sets is provably necessary: Maehara \cite{maehara2015risk} proved that restricted to single sets, it is NP-hard to compute any multiplicative approximation to the CVaR of a submodular set function.
	
	In this overview, we focus on the continuous setting; details on the reduction from the discrete portfolio problem to continuous submodular optimization can be found in the full paper \cite{wilder2018risk}. Our main contribution is the RASCAL algorithm, which computes a $(1 - 1/e)$-approximation to optimizing the CVaR of a smooth, continuous submodular function (up to an additive loss of $\epsilon$). RASCAL jointly exploits properties of both submodularity and the CVaR to provably approximate the non-concave maximization problem. We start out by formalizing the problem.

	\textbf{Continuous submodularity: }Let $\mathcal{X} = \prod_{i = 1}^n \mathcal{X}_i$ be a subset of $R^n$, where each $\mathcal{X}_i$ is a compact subset of $R$. A twice-differentiable function $F: \mathcal{X} \to R$ is \emph{diminishing returns submodular} (DR-submodular) if for all $\bm{x} \in \mathcal{X}$ and all $i,j = 1...n$, $\frac{\partial^2 F(\bm{x})}{\partial x_i \partial x_j} \leq 0$ \cite{bian2017guaranteed}. Intuitively, the gradient of $F$ only shrinks as $\bm{x}$ grows, just as the marginal gains of a submodular set function only decrease as items are added. Continuous submodular functions need not be convex or concave (concavity requires that the Hessian is negative semi-definite, not that the individual entries are nonpositive). We consider \emph{monotone} functions, where $F(\bm{x}) \leq F(\bm{y}) \,\,\forall \bm{x} \preceq \bm{y}$ ($\preceq$ denotes element-wise inequality). We assume that $F$ lies in $[0,M]$ for some constant $M$. Without loss of generality, we assume $F(0) = 0$ (normalization).
	
	In our setting $F$ is a function of both the decision variables $\bm{x}$ and a random parameter $y$. Specifically, we consider functions $F(\bm{x}, y)$ where $F(\cdot, y)$ is continuous submodular in $\bm{x}$ for each fixed $y$. We allow any DR-submodular $F$ which satisfies some standard smoothness conditions. First, we assume that $F$ is $L_1$-Lipschitz for some constant $L_1$ (for concreteness, with respect to the $\ell_2$ norm\footnote{We use the $\ell_2$ norm for concreteness. However, our arguments easily generalize to any $\ell_p$ norm.}). Second, we assume that $F$ is twice differentiable with $L_2$-Lipschitz gradient. Third, we assume that $F$ has bounded gradients, $||\nabla F||_2 \leq G$. Only the last condition is strictly necessary; our approach can be extended to any $F$ with bounded gradients via known techniques \cite{duchi2012randomized}. 
	
	\textbf{Conditional value at risk: } Intuitively, the CVaR measures performance in the $\alpha$ worst fraction of cases. First, we define the \emph{value at risk} at level $\alpha \in [0,1]$:
	
	\begin{align*}
	\text{VaR}_\alpha(\bm{x}) = \inf \{ \tau \in R : \text{Pr}_y\left[F(\bm{x}, y) \leq \tau\right] \geq \alpha \}.
	\end{align*}
	
	That is, $\text{VaR}_\alpha(\bm{x})$ is the $\alpha$-quantile of the random variable $F(\bm{x}, y)$. CVaR is the expectation of $F(\bm{x}, y)$, conditioned on it falling into this set of $\alpha$-worst cases:
	
	\begin{align*}
	\text{CVaR}_\alpha(\bm{x}) = \E_y\left[F(\bm{x}, y) | F(\bm{x}, y) \leq \text{VaR}_\alpha (\bm{x})\right].
	\end{align*}
	
	CVaR is a more popular risk measure than VaR both because it counts the impact of the entire $\alpha$-tail of the distribution and because it has better mathematical properties \cite{rockafellar2000optimization}.
	
	\textbf{Optimization problem: } We consider the problem of maximizing $\cvar(\bm{x})$ over $\bm{x}$ belonging to some feasible set $\mathcal{P}$. We allow $\mathcal{P}$ to be any downward closed polytope. A polytope is downward closed if there is a lower bound $\bm{\ell}$ such that $\bm{x} \succeq \bm{\ell} \,\, \forall \bm{x} \in \mathcal{P}$ and for any $\bm{y} \in \mathcal{P}$, $\bm{\ell} \preceq \bm{x} \preceq \bm{y}$ implies that $\bm{x} \in \mathcal{P}$.  Without loss of generality, we assume that $\mathcal{P}$ is entirely nonnegative with $\bm{\ell} = 0$. Otherwise, we can define the translated set $\mathcal{P}' = \{\bm{x} - \bm{\ell} : \bm{x} \in \mathcal{P}\}$ and corresponding function $F'(\bm{x}, y) = F(\bm{x} - \bm{\ell}, y)$. Let $d = \max_{\bm{x}, \bm{y} \in \mathcal{P}} ||\bm{x} - \bm{y}||_2$ be the diameter of $\mathcal{P}$.
	
	We want to solve the problem $\max_{\bm{x} \in \mathcal{P}} \cvar(\bm{x})$. It is important to note that $\cvar(\bm{x})$ need \emph{not} be a smooth DR-submodular function in $\bm{x}$. However, we would like to leverage the nice properties of the underling $F$. Towards this end, we note that the above problem can be rewritten in a more useful form \cite{rockafellar2000optimization}. Let $[t]^+ = \max(t, 0)$. Maximizing $\cvar(\bm{x})$ is equivalent to solving
	
	\begin{align}
	\max_{\bm{x} \in \mathcal{P}, \tau \in [0,M]} H(\bm{x},\tau) = \tau - \frac{1}{\alpha} \E\left[\left[\tau - F(\bm{x}, y)\right]^+\right]\label{problem:cvar}
	\end{align}
	
	where $\tau$ is an auxiliary parameter. For any fixed $\bm{x}$, the optimal value of $\tau$ is $\var(\bm{x})$ \cite{rockafellar2000optimization}. It is known that when $F(\cdot, y)$ is \emph{concave} in $\bm{x}$, this is a concave optimization problem. However, little is known when $F$ may be nonconcave.
	
	We now introduce the RASCAL (Risk Averse Submodular optimization via Conditional vALue at risk) algorithm for continuous submodular CVaR optimization. 
	RASCAL solves Problem \ref{problem:cvar}, which is a function of both the decision variables $\bm{x}$ and the auxiliary parameter $\tau$. Roughly, $\tau$ should be understood as a threshold maintained by the algorithm for what constitutes a ``bad" scenario: at each iteration, RASCAL tries to increase $F(\bm{x}, y)$ for those scenarios $y$ such that $F(\bm{x}, y) \leq \tau$. 
	
	More formally, RASCAL is a coordinate ascend style algorithm. Each iteration first makes a Frank-Wolfe style update to $\bm{x}$. Recall that Frank-Wolfe is a gradient-based algorithm originally developed for concave optimization. However, it can be modified to maximize continuous submodular functions \cite{bian2017guaranteed}. RASCAL then sets $\tau$ to its optimal value given the current $\bm{x}$. This approach is motivated by the unique properties of the CVaR objective $H$. It can be shown that $H$ is jointly up-concave in the variable $(\bm{x}, \tau)$. However, $H$ is not monotone in $\tau$. Indeed, $H$ is decreasing in $\tau$ for $\tau > \text{VaR}_\alpha(\bm{x})$. The Frank-Wolfe algorithm relies crucially on monotonicity; nonmonotonicity is much more difficult to handle. 
	
	Instead, we exploit a unique form of structure in $H$. Specifically, $H$ is monotone in $\bm{x}$, but only up-concave (not fully concave). Conversely, while $H$ is nonmonotone in $\tau$, we can easily solve the one-dimensional problem $\max_{\tau \in [0, M]} H(\bm{x}, \tau)$ for any fixed $\bm{x}$ (see the full paper for details). Our approach makes use of both properties: the Frank-Wolfe update leverages monotone up-concavity in $\bm{x}$, while the update to $\tau$ leverages easy solvability of the one-dimensional subproblem. 
	
	In order to make this approach work, two ingredients are necessary. First, we need access to the gradient of $H$ in order to implement the Frank-Wolfe update for $\bm{x}$. Unfortunately, $H$ is not even differentiable everywhere. We instead present a smoothed estimator $\textsc{SmoothGrad}$ which restores differentiability at the cost of introducing a controlled amount of bias. Second, we need to solve the one-dimensional problem of finding the optimal value of $\tau$. We in fact introduce a subroutine $\textsc{SmoothTau}$ which solves a smoothed version of the optimal $\tau$ problem. In the end, we obtain the following theoretical guarantee:
	
	\begin{theorem}
		For any $\epsilon > 0$, RASCAL outputs a solution $\bm{x} \in \mathcal{P}$ satisfying $\cvar(\bm{x}) \geq (1 - 1/e)OPT - \epsilon$ with probability at least $1 - \delta$. There are $K = O\left(\frac{L_2 d^2}{\alpha \epsilon} + \frac{L_1 G d^2}{\alpha^2 \epsilon^2}\right)$ iterations, requiring $O\left(sK\right)$ total evaluations of $F$, $O\left(sK\right)$ evaluations of $\nabla F$, and $K$ calls to a linear optimization oracle for $\mathcal{P}$. Here, $s = O\left(\frac{n M^2}{\epsilon^2}\log \frac{1}{\delta} \log \frac{L_1}{\epsilon}\right)$ is the number of samples taken from the underlying distribution.  \label{theorem:cvar}
	\end{theorem}
	
	\section{Infectious disease prevention} \label{section:disease}

	Treatable infectious diseases cause hundreds of thousands of cases of disability and death worldwide. Often, this burden is caused by long-term diseases which are continuously present in the population, as opposed to short-term epidemics like influenza. For instance, tuberculosis (TB) deaths in India numbered over 480,000 in 2014 \cite{WHO2015}, and even developed nations like the U.S.\ have observed over 395,000 cases of gonorrhea in 2015 \cite{CDC}. In both cases, many individuals remain undiagnosed although treatment is available. Outreach efforts to increase screening can lower disease burden; e.g., the Indian government conducts advertising campaigns for TB awareness. Limited resources require these campaigns to be carefully targeted at the most effective groups for reducing disease. Targeting is complicated by changing population dynamics, as individuals age and migrate over time, as well as by uncertainty around disease transmission rates. Officials currently make such decisions by hand as no algorithmic assistance is available.
	
	To remedy this situation, we design an algorithm to divide a limited outreach budget between demographic groups in order to minimize long term disease prevalence under uncertain population dynamics. Our approach contrasts with existing algorithms for disease control, which often consider disease spread between nodes on a static graph \cite{saha2015approximation,borgs2010distribute}. This is a sensible model of short term disease spread but is less suitable for long-term planning in diseases such as TB or gonorrhea, where people are born, die, age, and move \cite{luke2012systems}. Accounting for changes in the underlying agents is particularly salient for a policymaker who must divide resources between demographic groups over many years to maximize societal long-term health. For instance, India produces 5 year plans to combat TB \cite{rntcpProgress}. Our approach also contrasts with previous work on agent-based disease models \cite{jindal2017agent,lee2010computer}. Such models may include realistic behaviors, but their complexity usually precludes algorithmic approaches to finding the optimal policy in an entire feasible set.
	
	An additional challenge, largely unexplored in previous algorithmic work, is that of uncertainty. Data is always limited; policymakers are never sure of exactly how many people are infected in each group, or of the contact patterns between them. In order to impact real world policy, algorithms for resource allocation must account for such uncertainties.
	
	We introduce a model which both captures underlying agent dynamics and can be solved using an algorithmic approach in a stochastic setting.  We make four main contributions, which are explored in detail in the full paper \cite{wilder2018preventing}. \emph{First}, we present the MCF-SIS model (Multiagent Continuous Flow-SIS) where disease spreads in a multiagent system with birth, death, and movement. The system evolves according to SIS (susceptible-infected-susceptible) dynamics and is stratified across age groups. This introduces a new problem in multiagent systems: computing the optimal resource allocation under MFS-SIS, as in the case where an outreach campaign must decide how to divide limited advertising dollars (or rupees) between the groups. 
	
	MCF-SIS introduces a continuous, nonconvex, highly nonlinear optimization problem which cannot be solved by existing methods. Many factors must be accounted for.  E.g., between-group disease transmission makes focusing on the groups with the most infected agents suboptimal. Moreover, agents in a targeted group are not cured instantaneously, so, e.g., to reduce prevalence in age group 30, we may need to start targeting resources at age 27. Lastly, we consider a stochastic setting where parts of the model (contact patterns between agents, the number of infected agents in each group, etc.) are not known exactly but are drawn from a distribution.

	Our \emph{second} contribution shows that optimal allocation in MCF-SIS is a \emph{continuous submodular} problem. This opens up a novel set of optimization techniques which have not previously been used in disease prevention. Continuous submodularity generalizes submodular set functions to continuous domains. Intuitively, infections averted by spending one unit of treatment resources can no longer be averted by additional spending, creating diminishing returns. \emph{Continuous submodularity is deliberately enabled by our modeling choices, in particular our shift from the discrete, graph-based setting common in previous work \cite{saha2015approximation,borgs2010distribute} to a continuous, population-based model.}
	
	Our \emph{third} contribution is a new algorithm called DOMO (Disease Outreach via Multiagent Optimization), which obtains an efficient $(1 - 1/e)$-approximation to the optimal allocation. Our algorithm builds on a recent theoretical framework for submodular optimization \cite{bian2017guaranteed}. DOMO's generalization of this framework to the stochastic setting may be of independent interest. 
	
	Our \emph{fourth} contribution is to instantiate MCF-SIS in two domains using empirical data which takes into account behavioral, demographic, and epidemic trends: first, TB spread in India, and second, gonorrhea in the United States. DOMO averts 8,000 annual person-years of TB and 20,000 person-years of gonorrhea compared to current policy.

	\section{Future directions} \label{section:future}
	
	There are many promising future questions related to algorithmic social intervention. Here, I detail two directions in progress.
	
	\subsection{Multi-fidelity optimization}
	
	Oftentimes, finding the best intervention amounts to performing optimization in a complex model. For instance, epidemiologists have build enormously complicated models of disease spread, which simulate the (stochastic) interactions of millions of agents and account for a range of factors. Such models are very faithful to what are believed to be the real-world processes of diseases spread, but suffer from very high computational cost and are poorly understood from an optimization perspective. Hence, researchers interested in optimization (including my work \cite{wilder2018preventing}) seek simpler and more tractable models. It is hoped that these simpler models are sufficiently faithful to reality to yield useful insights, but they will clearly not be as accurate as more detailed simulations. 
	
	Hence, a natural direction is to pursue better methods for \emph{multi-fidelity} optimization: using a simpler model as a guide, or surrogate, to optimize a more complex one. Such methods have recently attracted interest in machine learning for use in hyperparameter optimization \cite{kandasamy2016gaussian}, and have previously been studied in several engineering disciplines \cite{sun2010two,forrester2007multi,alexandrov2001approximation}. However, previous models suffer from a variety of shortcomings in how they treat both the high-fidelity and surrogate models. For instance, most do not incorporate stochasticity in the high-fidelity model, where only noisy observations of the ground truth are available. This can easily become problematic because randomness is ubiquitous in modeling, especially in noisy domains like human interaction. With respect to the low-fidelity model, previous work usually assumes black-box access. However, this neglects the potential advantage that can be gained through exploiting known structure in the surrogate. For instance, if we were to use the MCF-SIS model introduced in \cite{wilder2018preventing} as a surrogate for a complex disease model, the DOMO algorithm can be used to find provably good approximate solutions. 
	
	Accordingly, the purpose of this project is to remedy such shortcomings by proposing multi-fidelity optimization methods which naturally incorporate stochasticity and leverage known structure in the surrogate model. The immediate application for such techniques is optimizing policies for preventing disease spread, but many other application areas are possible.
	
	\subsection{Distributionally robust submodular optimization}
	
	In this project, we consider a more nuanced treatment of uncertainty in submodular optimization, which yields improved properties in learning and optimizing from limited data. Suppose that we wish to maximize a submodular function which is not known exactly. For instance, we may have a finite collection of samples from an unknown distribution and wish to maximize expected performance over that distribution. Or, we may have a probabilistic model (e.g., a model of influence spread) but do not believe that this model is exactly correct. We can draw samples from this model and optimize empirical performance over the samples (the de facto approach in influence maximization), but such a process will not incorporate our uncertainty about the true distribution that the objective is drawn from. 
	
	In both settings (limited data and model uncertainty), is there a better approach than maximizing empirical performance on the samples? One attractive alternative is \emph{distributionally robust} optimization. Let the empirical distribution on sample objective functions $f_1...f_n$ be denoted by $\hat{p}_n$. Let $D(p||\hat{p}_n)$ be a divergence measure between another distribution $p$ and the empirical distribution $\hat{p}_n$ (e.g., the $\chi^2$ divergence). The distributionally robust optimization problem is to solve
	
	\begin{align*}
	\max_S \min_{p: D(p||\hat{p}_n) \leq \rho} \E_{f \sim p}[f(S)].
	\end{align*}
	
	That is, we aim to maximize our worst-case expected performance over all distributions that are ``close" to the observed distribution $\hat{p}_n$. One advantage of this formulation is that it can be seen as maximizing a high-probability bound on expected performance. Let $\mathcal{D}$ be the unknown distribution generating the objective. Given $n$ samples from $\mathcal{D}$, classical arguments (e.g., the Bernstsein bound) show that 
	
	\begin{align*}
	\E_{f \sim \mathcal{D}}[f(S)] \geq \E_{f \sim \hat{p}_n}[f(S)] - C_1 \sqrt{\frac{\text{Var}_{\mathcal{D}}[f(S)]}{n}}
	\end{align*}
	
	where $C_1$ is a constant (e.g., depending on the probability with which we want the bound to hold). Hence, when the variance is large, we can do better by optimizing the entire term on the right-hand side instead of just empirical performance on the samples (the first term). It has recently been shown \cite{namkoong2017variance} that (under some conditions) the distributionally robust problem corresponds exactly to such a variance-regularized objective. This has led to improved generalization for \emph{convex} loss functions, where distributionally robust optimization remains a convex optimization problem. The purpose of this project is to extend distributionally robust techniques to submodular optimization. This will entail the development of new algorithmic tools to deal with the (natively combinatorial) nonconvex problem. However, such development is a very relevant direction for algorithmic social intervention since objectives in many problems are inferred from limited data or uncertain models.

	\bibliographystyle{plain}
	\bibliography{proposal_bib}

\begin{thebibliography}{10}

\bibitem{abbe2015community}
Emmanuel Abbe and Colin Sandon.
\newblock Community detection in general stochastic block models: Fundamental
  limits and efficient algorithms for recovery.
\newblock In {\em FOCS}, pages 670--688. IEEE, 2015.

\bibitem{alexandrov2001approximation}
Natalia~M Alexandrov, Robert~Michael Lewis, Clyde~R Gumbert, Lawrence~L Green,
  and Perry~A Newman.
\newblock Approximation and model management in aerodynamic optimization with
  variable-fidelity models.
\newblock {\em Journal of Aircraft}, 38(6):1093--1101, 2001.

\bibitem{alon2012optimizing}
Noga Alon, Iftah Gamzu, and Moshe Tennenholtz.
\newblock Optimizing budget allocation among channels and influencers.
\newblock In {\em WWW}, pages 381--388, 2012.

\bibitem{bach2015submodular}
Francis Bach.
\newblock Submodular functions: from discrete to continous domains.
\newblock {\em arXiv preprint arXiv:1511.00394}, 2015.

\bibitem{bian2017guaranteed}
Andrew~An Bian, Baharan Mirzasoleiman, Joachim~M. Buhmann, and Andreas Krause.
\newblock Guaranteed non-convex optimization: Submodular maximization over
  continuous domains.
\newblock In {\em AISTATS}, 2017.

\bibitem{bond201261}
Robert~M Bond, Christopher~J Fariss, Jason~J Jones, Adam~DI Kramer, Cameron
  Marlow, Jaime~E Settle, and James~H Fowler.
\newblock A 61-million-person experiment in social influence and political
  mobilization.
\newblock {\em Nature}, 489(7415):295--298, 2012.

\bibitem{borgs2010distribute}
Christian Borgs, Jennifer Chayes, Ayalvadi Ganesh, and Amin Saberi.
\newblock How to distribute antidote to control epidemics.
\newblock {\em Random Structures \& Algorithms}, 37(2):204--222, 2010.

\bibitem{bosansky2014exact}
Branislav Bosansky, Christopher Kiekintveld, Viliam Lisy, and Michal Pechoucek.
\newblock An exact double-oracle algorithm for zero-sum extensive-form games
  with imperfect information.
\newblock {\em Journal of Artificial Intelligence Research}, 51:829--866, 2014.

\bibitem{calinescu2011maximizing}
Gruia Calinescu, Chandra Chekuri, Martin P{\'a}l, and Jan Vondr{\'a}k.
\newblock Maximizing a monotone submodular function subject to a matroid
  constraint.
\newblock {\em SIAM Journal on Computing}, 40(6):1740--1766, 2011.

\bibitem{CDC}
CDC.
\newblock {Reported STDs in the United States.}, 2015.

\bibitem{chekuri2010dependent}
Chandra Chekuri, Jan Vondrak, and Rico Zenklusen.
\newblock Dependent randomized rounding via exchange properties of
  combinatorial structures.
\newblock In {\em FOCS}, 2010.

\bibitem{chen2017robust}
Robert Chen, Brendan Lucier, Yaron Singer, and Vasilis Syrgkanis.
\newblock Robust optimization for non-convex objectives.
\newblock In {\em NIPS}, 2017.

\bibitem{chen2010scalable}
Wei Chen, Chi Wang, and Yajun Wang.
\newblock Scalable influence maximization for prevalent viral marketing in
  large-scale social networks.
\newblock In {\em KDD}, pages 1029--1038. ACM, 2010.

\bibitem{chen2014cim}
Yi-Cheng Chen, Wen-Yuan Zhu, Wen-Chih Peng, Wang-Chien Lee, and Suh-Yin Lee.
\newblock Cim: community-based influence maximization in social networks.
\newblock {\em ACM Transactions on Intelligent Systems and Technology (TIST)},
  5(2):25, 2014.

\bibitem{duchi2012randomized}
John~C Duchi, Peter~L Bartlett, and Martin~J Wainwright.
\newblock Randomized smoothing for stochastic optimization.
\newblock {\em SIAM Journal on Optimization}, 22(2):674--701, 2012.

\bibitem{fienberg1981categorical}
Stephen~E Fienberg and Stanley~S Wasserman.
\newblock Categorical data analysis of single sociometric relations.
\newblock {\em Sociological methodology}, 12:156--192, 1981.

\bibitem{forrester2007multi}
Alexander~IJ Forrester, Andr{\'a}s S{\'o}bester, and Andy~J Keane.
\newblock Multi-fidelity optimization via surrogate modelling.
\newblock In {\em Proceedings of the royal society of london a: mathematical,
  physical and engineering sciences}, volume 463, pages 3251--3269. The Royal
  Society, 2007.

\bibitem{jain2013security}
Manish Jain, Vincent Conitzer, and Milind Tambe.
\newblock Security scheduling for real-world networks.
\newblock In {\em AAMAS}, 2013.

\bibitem{jain2011double}
Manish Jain, Dmytro Korzhyk, Ond{\v{r}}ej Van{\v{e}}k, Vincent Conitzer, Michal
  P{\v{e}}chou{\v{c}}ek, and Milind Tambe.
\newblock A double oracle algorithm for zero-sum security games on graphs.
\newblock In {\em AAMAS}, 2011.

\bibitem{jindal2017agent}
Akshay Jindal and Shrisha Rao.
\newblock Agent-based modeling and simulation of mosquito-borne disease
  transmission.
\newblock In {\em Proceedings of the 16th Conference on Autonomous Agents and
  MultiAgent Systems}, pages 426--435. International Foundation for Autonomous
  Agents and Multiagent Systems, 2017.

\bibitem{jung2012irie}
Kyomin Jung, Wooram Heo, and Wei Chen.
\newblock Irie: Scalable and robust influence maximization in social networks.
\newblock In {\em ICDM}, pages 918--923. IEEE, 2012.

\bibitem{kandasamy2016gaussian}
Kirthevasan Kandasamy, Gautam Dasarathy, Junier~B Oliva, Jeff Schneider, and
  Barnab{\'a}s P{\'o}czos.
\newblock Gaussian process bandit optimisation with multi-fidelity evaluations.
\newblock In {\em Advances in Neural Information Processing Systems}, pages
  992--1000, 2016.

\bibitem{kempe2003maximizing}
David Kempe, Jon Kleinberg, and {\'E}va Tardos.
\newblock Maximizing the spread of influence through a social network.
\newblock In {\em KDD}, 2003.

\bibitem{kocckesen2000strategic}
Levent Ko{\c{c}}kesen, Efe~A Ok, and Rajiv Sethi.
\newblock The strategic advantage of negatively interdependent preferences.
\newblock {\em Journal of Economic Theory}, 92(2):274--299, 2000.

\bibitem{krause2008robust}
Andreas Krause, H~Brendan McMahan, Carlos Guestrin, and Anupam Gupta.
\newblock Robust submodular observation selection.
\newblock {\em Journal of Machine Learning Research}, 9(Dec):2761--2801, 2008.

\bibitem{krause2011randomized}
Andreas Krause, Alex Roper, and Daniel Golovin.
\newblock Randomized sensing in adversarial environments.
\newblock In {\em IJCAI}, 2011.

\bibitem{krzakala2013spectral}
Florent Krzakala, Cristopher Moore, Elchanan Mossel, Joe Neeman, Allan Sly,
  Lenka Zdeborov{\'a}, and Pan Zhang.
\newblock Spectral redemption in clustering sparse networks.
\newblock {\em PNAS}, 110(52):20935--20940, 2013.

\bibitem{kulesza2012determinantal}
Alex Kulesza and Ben Taskar.
\newblock Determinantal point processes for machine learning.
\newblock {\em Foundations and Trends in Machine Learning}, 5(2--3):123--286,
  2012.

\bibitem{lee2010computer}
Bruce~Y Lee, Shawn~T Brown, George~W Korch, Philip~C Cooley, Richard~K
  Zimmerman, William~D Wheaton, Shanta~M Zimmer, John~J Grefenstette, Rachel~R
  Bailey, Tina-Marie Assi, et~al.
\newblock A computer simulation of vaccine prioritization, allocation, and
  rationing during the 2009 h1n1 influenza pandemic.
\newblock {\em Vaccine}, 28(31):4875--4879, 2010.

\bibitem{leskovec2009community}
Jure Leskovec, Kevin~J Lang, Anirban Dasgupta, and Michael~W Mahoney.
\newblock Community structure in large networks: Natural cluster sizes and the
  absence of large well-defined clusters.
\newblock {\em Internet Mathematics}, 6(1):29--123, 2009.

\bibitem{luke2012systems}
Douglas~A Luke and Katherine~A Stamatakis.
\newblock Systems science methods in public health: dynamics, networks, and
  agents.
\newblock {\em Annual review of public health}, 33:357--376, 2012.

\bibitem{maehara2015risk}
Takanori Maehara.
\newblock Risk averse submodular utility maximization.
\newblock {\em Operations Research Letters}, 43(5):526--529, 2015.

\bibitem{mossel2015reconstruction}
Elchanan Mossel, Joe Neeman, and Allan Sly.
\newblock Reconstruction and estimation in the planted partition model.
\newblock {\em Probability Theory and Related Fields}, 162(3-4):431--461, 2015.

\bibitem{namkoong2017variance}
Hongseok Namkoong and John~C Duchi.
\newblock Variance-based regularization with convex objectives.
\newblock In {\em Advances in Neural Information Processing Systems}, pages
  2975--2984, 2017.

\bibitem{ohsaka2017portfolio}
Naoto Ohsaka and Yuichi Yoshida.
\newblock Portfolio optimization for influence spread.
\newblock In {\em WWW}, pages 977--985, 2017.

\bibitem{rice2012mobilizing}
Eric Rice, Eve Tulbert, Julie Cederbaum, Anamika~Barman Adhikari, and
  Norweeta~G Milburn.
\newblock Mobilizing homeless youth for {HIV} prevention.
\newblock {\em Health education research}, 27(2):226--236, 2012.

\bibitem{rntcpProgress}
RNTCP.
\newblock Revised national tuberculosis control programme annual status report.
\newblock New Delhi, India: Ministry of Health and Family Welfare.
  \url{http://tbcindia.nic.in/showfile.php?lid=3180}, 2016.

\bibitem{rockafellar2000optimization}
R~Tyrrell Rockafellar and Stanislav Uryasev.
\newblock Optimization of conditional value-at-risk.
\newblock {\em Journal of risk}, 2:21--42, 2000.

\bibitem{saha2015approximation}
Sudip Saha, Abhijin Adiga, B~Aditya Prakash, and Anil Kumar~S Vullikanti.
\newblock Approximation algorithms for reducing the spectral radius to control
  epidemic spread.
\newblock In {\em Proceedings of the SIAM International Conference on Data
  Mining}, pages 568--576. SIAM, 2015.

\bibitem{sampson2016assignment}
Thomas Sampson.
\newblock Assignment reversals: Trade, skill allocation and wage inequality.
\newblock {\em J. Econ. Theory}, 163:365--409, 2016.

\bibitem{soma2014optimal}
Tasuku Soma, Naonori Kakimura, Kazuhiro Inaba, and Ken-ichi Kawarabayashi.
\newblock Optimal budget allocation: Theoretical guarantee and efficient
  algorithm.
\newblock In {\em ICML}, pages 351--359, 2014.

\bibitem{staib2017robust}
Matthew Staib and Stefanie Jegelka.
\newblock Robust budget allocation via continuous submodular functions.
\newblock In {\em ICML}, 2017.

\bibitem{sun2010two}
Guangyong Sun, Guangyao Li, Michael Stone, and Qing Li.
\newblock A two-stage multi-fidelity optimization procedure for honeycomb-type
  cellular materials.
\newblock {\em Computational Materials Science}, 49(3):500--511, 2010.

\bibitem{tang2014influence}
Youze Tang, Xiaokui Xiao, and Yanchen Shi.
\newblock Influence maximization: Near-optimal time complexity meets practical
  efficiency.
\newblock In {\em KDD}. ACM, 2014.

\bibitem{valente2007identifying}
Thomas~W Valente and Patchareeya Pumpuang.
\newblock Identifying opinion leaders to promote behavior change.
\newblock {\em Health Education \& Behavior}, 2007.

\bibitem{vondrak2008optimal}
Jan Vondr{\'a}k.
\newblock Optimal approximation for the submodular welfare problem in the value
  oracle model.
\newblock In {\em STOC}, pages 67--74, 2008.

\bibitem{wang2010community}
Yu~Wang, Gao Cong, Guojie Song, and Kunqing Xie.
\newblock Community-based greedy algorithm for mining top-k influential nodes
  in mobile social networks.
\newblock In {\em KDD}, pages 1039--1048. ACM, 2010.

\bibitem{WHO2015}
WHO.
\newblock {World Health Organization. Tuberculosis country profiles.}
\newblock \url {http://www.who.int/tb/country/data/profiles/en/}, 2015.

\bibitem{wilder2018equilibrium}
Bryan Wilder.
\newblock Equilibrium computation and robust optimization in zero sum games
  with submodular structure.
\newblock In {\em AAAI}, 2018.

\bibitem{wilder2018risk}
Bryan Wilder.
\newblock Risk-sensitive submodular optimization.
\newblock In {\em Proceedings of the 32nd AAAI Conference on Artificial
  Intelligence}, 2018.

\bibitem{wilder2018maximizing}
Bryan Wilder, Nicole Immorlica, Eric Rice, and Milind Tambe.
\newblock Maximizing influence in an unknown social network.
\newblock In {\em AAAI}, 2018.

\bibitem{wilder2018end}
Bryan Wilder, Laura Onasch-Vera, Juliana Hudson, Jose Luna, Nicole Wilson,
  Robin Petering, Darlene Woo, Milind Tambe, and Eric Rice.
\newblock End-to-end influence maximization in the field, 2018.

\bibitem{wilder2018optimizing}
Bryan Wilder, Han~Ching Ou, Kayla de~la Haye, and Milind Tambe.
\newblock Optimizing network structure for preventative health.
\newblock In {\em AAMAS}, 2018.

\bibitem{wilder2018preventing}
Bryan Wilder, Sze-Chuan Suen, and Milind Tambe.
\newblock Preventing infectious disease in dynamic populations under
  uncertainty.
\newblock In {\em AAAI}, 2018.

\bibitem{wilder2017uncharted}
Bryan Wilder, Amulya Yadav, Nicole Immorlica, Eric Rice, and Milind Tambe.
\newblock Uncharted but not uninfluenced: Influence maximization with an
  uncertain network.
\newblock In {\em AAMAS}, 2017.

\bibitem{wilder2017unknown}
Bryan Wilder, Amulya Yadav, Nicole Immorlica, Eric Rice, and Milind Tambe.
\newblock Uncharted but not uninfluenced: Influence maximization with an
  uncertain network.
\newblock In {\em AAMAS}, pages 740--748, 2017.

\bibitem{yadav2016using}
Amulya Yadav, Hau Chan, Albert Xin~Jiang, Haifeng Xu, Eric Rice, and Milind
  Tambe.
\newblock Using social networks to aid homeless shelters: Dynamic influence
  maximization under uncertainty.
\newblock In {\em AAMAS}, pages 740--748, 2016.

\bibitem{yadav2017influence}
Amulya Yadav, Bryan Wilder, Eric Rice, Robin Petering, Jaih Craddock, Amanda
  Yoshioka-Maxwell, Mary Hemler, Laura Onasch-Vera, Milind Tambe, and Darlene
  Woo.
\newblock Influence maximization in the field: The arduous journey from
  emerging to deployed application.
\newblock In {\em AAMAS}, 2017.

\end{thebibliography}
\end{document}